\documentclass[sigconf]{acmart}
\AtBeginDocument{%
  }

\copyrightyear{2025}
\acmYear{2025}
\setcopyright{acmlicensed}
\acmConference[KDD '25]{Proceedings of the 31st ACM SIGKDD Conference on Knowledge Discovery and Data Mining V.2}{August 3--7, 2025}{Toronto, ON, Canada}
\acmBooktitle{Proceedings of the 31st ACM SIGKDD Conference on Knowledge Discovery and Data Mining V.2 (KDD '25), August 3--7, 2025, Toronto, ON, Canada}
\acmDOI{10.1145/3711896.3737015}
\acmISBN{979-8-4007-1454-2/25/08}

\usepackage{enumitem}
\usepackage{graphicx}
\usepackage{algorithm}
\usepackage[noend]{algpseudocode}
\usepackage{adjustbox}
\usepackage{multirow}
\usepackage{colortbl}
\usepackage{booktabs}
\usepackage{tikz}
\usepackage{pgfplots}
\usepackage{subfigure}
\pgfplotsset{compat=1.17}
\usepackage{caption}
\usepackage{listings}
\usepackage{tcolorbox}
\usepackage{balance}

\makeatletter
\renewcommand\@fnsymbol[1]{\ensuremath{\ifcase#1\or *\or \dagger\or \ddagger\or \mathsection\or \mathparagraph\or \|\or **\or \dagger\dagger\or \ddagger\ddagger\else\@ctrerr\fi}}
\makeatother

\begin{document}

\title{KnowTrace: Bootstrapping Iterative Retrieval-Augmented Generation with Structured Knowledge Tracing}

\author{Rui Li}
\authornote{Work done during Rui Li's internship at Huawei Noah's Ark Lab.}
\authornote{Beijing Key Laboratory of Research on Large Models and Intelligent Governance.}
\authornote{Engineering Research Center of Next-Generation Intelligent Search and Recommendation, MOE.}
\affiliation{
  \institution{\mbox{Gaoling School of Artificial Intelligence}, \mbox{Renmin University of China}}
  \city{Beijing}
  \country{China}
}
\email{lirui121200@ruc.edu.cn}

\author{Quanyu Dai}
\affiliation{
  \institution{Huawei Noah's Ark Lab}
  \city{Shenzhen}
  \country{China}}
\email{daiquanyu@huawei.com}

\author{Zeyu Zhang}
\authornotemark[2]
\authornotemark[3]
\affiliation{
  \institution{\mbox{Gaoling School of Artificial Intelligence}, \mbox{Renmin University of China}}
  \city{Beijing}
  \country{China}}
\email{zeyuzhang@ruc.edu.cn}

\author{Xu Chen}
\authornotemark[2]
\authornotemark[3]
\authornote{Corresponding author.}
\affiliation{
 \institution{\mbox{Gaoling School of Artificial Intelligence}, \mbox{Renmin University of China}}
  \city{Beijing}
  \country{China}}
\email{xu.chen@ruc.edu.cn}

\author{Zhenhua Dong}
\affiliation{
 \institution{Huawei Noah's Ark Lab}
  \city{Shenzhen}
  \country{China}}
\email{dongzhenhua@huawei.com}

\author{Ji-Rong Wen}
\authornotemark[2]
\authornotemark[3]
\affiliation{
 \institution{\mbox{Gaoling School of Artificial Intelligence}, \mbox{Renmin University of China}}
  \city{Beijing}
  \country{China}}
\email{jrwen@ruc.edu.cn}


\begin{abstract}
Recent advances in retrieval-augmented generation (RAG) furnish large language models (LLMs) with iterative retrievals of relevant information to handle complex multi-hop questions. These methods typically alternate between LLM reasoning and retrieval to accu\-mulate external information into the LLM's context. However, the \emph{ever-growing context} inherently imposes an increasing burden on the LLM to perceive connections among critical information pieces, with \emph{futile reasoning steps} further exacerbating this overload issue. In this paper, we present \textbf{KnowTrace}, an elegant RAG framework to (1) \emph{mitigate the context overload} and (2) \emph{bootstrap higher-quality multi-step reasoning}. Instead of simply piling the retrieved contents, KnowTrace autonomously traces out desired knowledge triplets to organize a specific knowledge graph relevant to the input question.
Such a structured workflow not only empowers the LLM with an intelligible context for inference, but also naturally inspires a reflective mechanism of \emph{knowledge backtracing} to identify contributive LLM generations as process supervision data for self-bootstrapping.
Extensive experiments show that KnowTrace consistently surpasses existing methods across three multi-hop question answering benchmarks, and the bootstrapped version further amplifies the gains.\footnote{The code is available at \url{https://github.com/rui9812/KnowTrace}.}
\end{abstract}




\maketitle

\section{Introduction}
Recent Large language models (LLMs)~\cite{llm1, dubey2024llama3} have shown impressive performance across a variety of natural language tasks through the form of question answering.
Despite their remarkable capabilities, LLMs continue to struggle with factual errors \cite{hallu1, hallu2, hallu3} when the input question exceeds their knowledge boundaries.
As a practical solution to this problem, 
Retrieval-Augmented Generation (RAG) \cite{rag1} empowers LLMs to incorporate external knowledge through information retrieval. 
One-time retrieval \cite{gao2023rag_survey, one-time3} usually suffices to fulfill the knowledge needs of single-hop questions, but
the complex \emph{multi-hop questions} still remain challenging due to their demands for intensive knowledge and multi-step reasoning 
capabilities, thus attracting widespread attention within the research community.

\begin{figure}[t!]
\centering
\includegraphics[width=0.80\columnwidth]{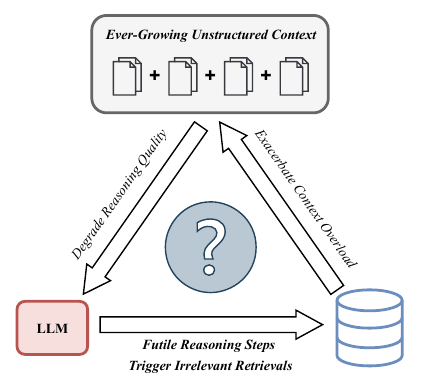}
\vspace{-2mm}
\caption{Two challenges of iterative RAG systems: ever-growing LLM context and non-contributive reasoning steps.} \label{fig:1}
\vspace{-4mm}
\end{figure}

To address the complex multi-hop questions, a range of recent works follow a natural strategy: extending the one-time RAG into a multi-round process \cite{IRCoT, Self-Ask, Iter-RetGen, react, IM-RAG}. 
Such iterative approaches interleave retrievals with LLM reasoning,
periodically incorporating new information to narrow semantic gaps between the multi-hop questions and their requisite knowledge \cite{Iter-RetGen}.
However, this work\-flow faces two critical challenges as presented in Figure \ref{fig:1}.
On the one hand, retrievers are not perfect and inherently inject massive redundant or misleading information into the context, imposing an increasing burden on the LLM to perceive the logical connections among critical information pieces, which are essential elements of reasoning \cite{relation}.
On the other hand, not every LLM reasoning step is contributive to the answer prediction (especially when confronted with an unintelligible context), and the futile reasoning generations would trigger the retrievals of irrelevant information, thus further exacerbating the issue of context overload.

Some one-time RAG works \cite{sg, infore, holmes, era-cot} ease the context burden with an auxiliary process of restructuring all retrieved documents, yet this process is impractical for the iterative RAG workflow, since it involves extensive LLM-driven operations (e.g., entity recognition and relation extraction) that would incur significant computational overhead for each retrieval iteration.
Moreover, several recent self-training techniques \cite{star, RFT, wei2025instructrag} focus on enhancing the single-step generation quality of LLMs, while how to bootstrap higher-quality multi-step reasoning capabilities for the iterative RAG systems still remains critical yet underexplored.

\begin{figure*}[t!]
\centering
\includegraphics[width=0.80\textwidth]{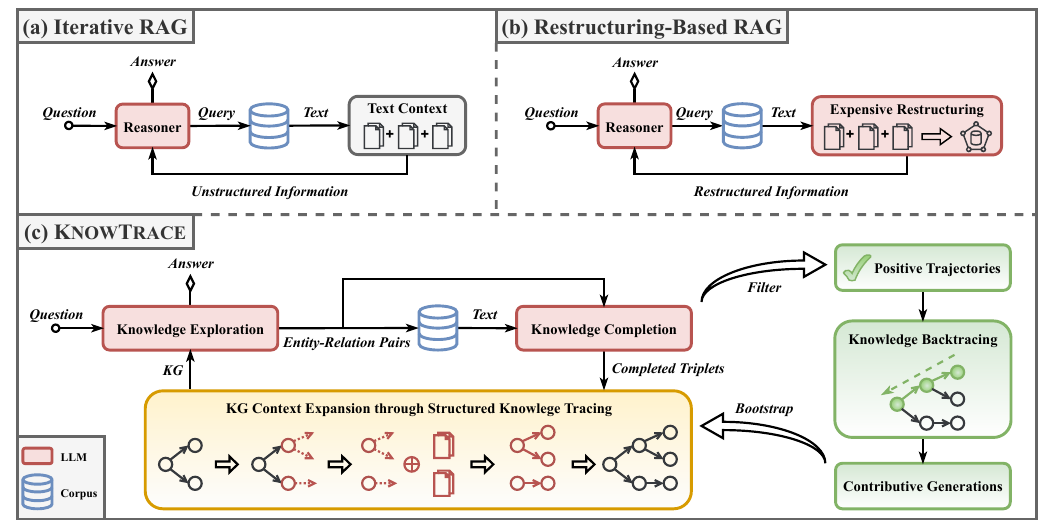}
\caption{An overview of two representative workflows (a-b) and our KnowTrace framework (c).} 
\label{fig:knowtrace}
\vspace{-2.5mm}
\end{figure*}

In this paper, we move beyond existing approaches by presenting a self-bootstrapping framework called \textbf{KnowTrace}, which adopts a unique perspective of \emph{structured knowledge tracing} to (1) tackle
the context overload issue and also (2) promote higher-quality multi-step reasoning in a self-taught manner.
Conceptually, we draw upon a profound insight from constructivist theory \cite{fosnot2013constructivism}: \emph{learning is never merely about accumulating information, but also involves proactively absorbing knowledge to construct and expand one's cognitive schema}.
Inspired by this principle, our framework reformulates the iterative RAG process as a coherent workflow of \emph{knowledge graph expansion}, as illustrated in Figure \ref{fig:knowtrace}(c).
Specifically, instead of plainly stacking or intricately restructuring all retrieved contents, KnowTrace treats the LLM as an active tracker to progressively \emph{explore-then-complete} question-relevant knowledge triplets, until tracing out a sufficient knowledge graph (KG) for the answer prediction.
Such an inference workflow seamlessly empowers the LLM with an intelligible context throughout the multi-step reasoning process, which clearly reveals critical knowledge structures and, as a result, inherently enhances the LLM's reasoning quality \cite{relation}.
Moreover, the transparency of KG expansion also spurs us to design a \emph{post-hoc backtracing mechanism}, allowing KnowTrace to \emph{retrospectively discern supportive knowledge triplets and contributive LLM generations based on the final inference}.
In this way, given the positive reasoning trajectories that ultimately lead to correct answers, our framework can automatically filter out the procedural impurities (i.e., non-contributive LLM generations) to synthesize high-quality process supervision data, thereby capable of elevating the multi-step reasoning capabilities via \emph{self-training}.
To sum up, the main contributions of this work are as follows:
\begin{itemize}[leftmargin=1.5em]
    \item We propose a self-bootstrapping iterative RAG framework called KnowTrace, which moves beyond existing methods by featuring a unique perspective of \emph{structured knowledge tracing}.
    \item During inference time, KnowTrace progressively traces question-relevant knowledge triplets to endow the LLM with \emph{an intelligible KG context}, thereby inherently raising the reasoning quality.
    \item By virtue of the self-organized KG structures, we further design a reflective mechanism of \emph{knowledge backtracing}, which enables KnowTrace to retrospectively synthesize higher-quality process supervision for effective self-taught finetuning.
    \item We conduct extensive experiments on three standard multi-hop question answering datasets. Under different configurations of LLMs and retrieval models, KnowTrace consistently surpasses current RAG methods across all the datasets. The self-elevated version (finetuned on the self-synthesized process supervision) further amplifies the performance advantages by a clear margin.
\end{itemize}

\newcommand{\para}[1]{\vspace{5pt} \noindent \textbf{#1}}
\section{Related Work}

This section reviews a broad range of relevant literature to position our proposed framework within the current research landscape.
More discussions are included in Appendix \ref{more_discussion}.

\para{Multi-Hop Question Answering (MHQA).} This task involves addressing multi-hop questions that require extensive knowledge and multi-step reasoning capabilities to comprehensively infer the final answers \cite{HotpotQA, 2Wiki, trivedi2022musique}. 
Different from the traditional approaches \cite{perez2020unsupervised,qi2020answering,deng2022interpretable}, this paper focuses on incorporating LLM reasoning with information retrieval to reason about complex multi-hop questions, which aligns with the recent RAG researches \cite{IRCoT,Self-Ask,Iter-RetGen,sg,sure,era-cot}.

\para{Iterative RAG Methods.} Retrieval-augmented generation (RAG) has been demonstrated as a promising technique to effectively boost the performance of LLMs in knowledge-intensive NLP tasks \cite{rag1}. 
Early RAG approaches \cite{zhu2021retrieving-then-read,early_rag_1,gao2023rag_survey} only perform one-time retrieval, struggling to gather all essential information for the input questions, especially for the complex multi-hop questions.
To ease this issue, a new series of iterative RAG methods have recently been developed \cite{Self-Ask, IRCoT, react, Iter-RetGen}.
These methods typically follow such a paradigm \cite{template} as shown in Figure \ref{fig:knowtrace}(a): first perform LLM reasoning to generate new queries (e.g., sub-questions) for retrieval, then accumulate the newly retrieved passages along with the previously collected ones to facilitate subsequent LLM reasoning, iteratively continuing this reasoning–retrieval process until the LLM derives the final answer.
Despite their prowess, these iterative methods inherently disregard the significance of the underlying knowledge structures behind the retrieved passages, which are essential elements of reasoning \cite{relation}.

\para{Structure-Enhanced RAG Methods.} In light of the limitation of purely unstructured information accumulation, some latest works follow a restructuring-based paradigm as shown in Figure \ref{fig:knowtrace}(b): they employ an auxiliary restructuring process for all retrieved passages \cite{sg,holmes,era-cot,infore} or even entire external corpora \cite{GraphRAG, sarmah2024hybridrag, GraphRAGSurvey}.
However, this strategy inherently necessitates extensive LLM invocations for intricate restructuring operations such as concept recognition and refinement, leading to significant computational costs and potential knowledge loss. 
In contrast, \emph{our KnowTrace framework seamlessly integrates structuring and reasoning into a coherent process, naturally enabling the acquisition of desired knowledge triplets at a low cost.}
In addition, several other structure-enhanced works \cite{[3],[4]} focus on parsing the multi-hop questions into masked structured chains, where each masked triplet is then either directly completed based on an existing KG or rewritten as a natural language query to access relevant passages from a text database.
However, these approaches heavily count on the accuracy of the initial parsing---errors at this stage can propagate---thereby requiring careful filtering operations and consistency checks \cite{[4]}.
Unlike them, \emph{our KnowTrace features a more flexible workflow of adaptively tracing relevant knowledge throughout the entire multi-step reasoning process, which effectively improves both performance and robustness.}

\para{Self-Taught Finetuning.} This technique refers to an impressive way to enhance the reasoning capabilities of LLMs by training them on their self-generated correct solutions \cite{star,RFT,rest,v-star}.
Specifically, this self-bootstrapping process \cite{star,v-star} corresponds to a simple loop: employ an LLM to infer a set of questions; collect all the generations that yield correct answers into a dataset; finetune the base LLM on this self-generated dataset; restart to collect new generations with the newly finetuned LLM. 
This self-taught process is founded upon such a priori assumption: the LLM generations that lead to correct answers reveal high-quality reasoning rationales. However, in the complex scenarios of MHQA, a long-horizon multi-step reasoning trajectory, even if it eventually leads to the correct answer, typically still contains irrelevant LLM generations, which would impair the effectiveness of subsequent finetuning.
\emph{Our KnowTrace framework is endowed with a reflective backtracing mechanism to filter out these procedural impurities, offering an elegant way to self-distill higher-quality process supervision data for the self-taught finetuning.}

\section{Methodology}
\subsection{Overview}
This work introduces KnowTrace, a new iterative RAG framework that can autonomously trace out the question-relevant KG contexts in a coherent manner to bootstrap the MHQA performance of LLMs. 
As shown in Figure \ref{fig:knowtrace}(c), KnowTrace alternately performs two LLM-driven operations: \emph{knowledge exploration} and \emph{knowledge completion}, to progressively acquire desired knowledge triplets during iterative retrievals until the expanding KG is sufficient for the LLM to output a definitive answer.
We detail this inference process in Section \ref{inf_sec}.
Moreover, by virtue of the transparent KG structures, KnowTrace
further leverages a backtracing mechanism to filter out useless LLM
generations (i.e., \emph{unavailing exploration} and \emph{extraneous completion}) from positive trajectories. In this way, KnowTrace can bootstrap its multi-step reasoning capabilities via finetuning on the self-distilled high-quality process supervision data, as described in Section \ref{sft_sec}.

\subsection{Structured Knowledge Tracing for Inference}
\label{inf_sec}

Given a multi-hop question $q$ and a textual corpus $C$, KnowTrace actively traces out a set of $q$-relevant knowledge triplets from $C$ to enrich an explicit KG context $\mathcal{G}_q=\{(e_s, r, e_o)|e_s,e_o\in\mathcal{E}_q,r\in\mathcal{R}_q\}$,
in which $\mathcal{E}_q$ and $\mathcal{R}_q$ denote the sets of entities and relationships, respectively, and each triplet $(e_s, r, e_o)$ reveals that there is a relation $r$ between subject entity $e_s$ and object entity $e_o$.
The entire inference workflow of KnowTrace corresponds to an iterative 
\emph{explore-then-complete} process (Algorithm \ref{alg:1}), where each iteration involves two LLM operations: \emph{knowledge exploration} and \emph{knowledge completion}.

\para{Knowledge Exploration.} 
During this phase, KnowTrace leverages the LLM's planning capability \cite{react} to determine the action for each iteration: either generate a definitive answer as the final prediction or continue to explore more relevant knowledge for KG expansion.
Formally, at the $l$-th iteration ($1\leq l\leq L$), KnowTrace integrates the question $q$ and the KG $\mathcal{G}_q^{l-1}$ acquired from the previous $l-1$ iterations (the initial $\mathcal{G}_q^{0}$ is empty) into an instruction prompt $I_{\tt exp}$.
This prompt is designed to elicit such a coherent generation from an LLM $M$: first, $M$ self-assesses whether $\mathcal{G}_q^{l-1}$ is sufficient to derive the final answer, and accordingly sets a boolean \texttt{FLAG}; if the \texttt{FLAG} is set to true, $M$ then directly outputs the answer $a$, along with a chain of thought $t$ \cite{CoT} to reveal the reasoning rationales; otherwise, $M$ then provides a specific exploration guidance on how to expand the current KG---it adaptively
determines the expansion points (i.e., entities) and the corresponding directions (i.e., relations), forming a set of entity-relation pairs that indicate the knowledge desired for the next reasoning step.
We formulate this process as follows:
\begin{equation}
\label{exp_func}
    \{\texttt{FLAG}, \mathcal{P}\}=M\left(I_{\tt exp}\left(q, \mathcal{G}_q^{l-1}\right)\right),
\end{equation}
where $\mathcal{P}$ is either the final prediction $[t,a]$ or the KG expansion guidance $\{(e_i,r_i)\}_{i=1}^P$ conditioned on
the self-generated \texttt{FLAG} as described above.
Note that $M$ can create new entities as the expansion points, rather than only selecting from the current KG. We refer to such entities as \emph{initial entities}, as they typically correspond to the expansion beginnings of different components.
After this phase, we then utilize each $(e_i,r_i)$ as the query to retrieve $N$ relevant passages from the textual corpus $C$, denoted as $\mathcal{C}^N_{(e_i,r_i)}$, while also employing this pair to guide the subsequent knowledge completion phase.

\newcommand{\algorithmicinput}{\textbf{Input:}}
\newcommand{\Input}{\item[\algorithmicinput]}
\newcommand{\algorithmicoutput}{\textbf{Output:}}
\newcommand{\Output}{\item[\algorithmicoutput]}

\begin{algorithm}[t!]
\caption{Inference Process of KnowTrace}\label{alg:1}
\begin{algorithmic}[1]
\Require base LLM $M$; prompts $I_{\tt exp}$ and $I_{\tt com}$; corpus $C$
\Input question $q$
\Output KG context $\mathcal{G}_q$; final thought $t$; final prediction $a$
\State $\mathcal{G}_q^0\leftarrow\emptyset$
\For{$l$ {\bfseries from} $1$ {\bfseries to} $L$}
    \State \textcolor{gray}{// Knowledge Exploration Phase in Equation (\ref{exp_func})}
    \State {$\{\texttt{FLAG}, \mathcal{P}\}\leftarrow M\left(I_{\tt exp}(q, \mathcal{G}_q^{l-1})\right)$}  
    \If{\texttt{FLAG}}
        \State \textcolor{gray}{// Chain-of-Thought Reasoning}
        \State $\mathcal{P}$ includes the thought $t$ and the prediction $a$
        \State \Return $\mathcal{G}_q^{l-1},t,a$
    \Else
        \State $\mathcal{P}$ includes a set of entity-relation pairs $\{(e_i,r_i)\}_{i=1}^P$
        \State \textcolor{gray}{// Parallelizable Inner Loop}
        \For{$i$ {\bfseries from} $1$ {\bfseries to} $P$}  
            \State $(e_i,r_i)$ serves as a query to retrieve $\mathcal{C}_{(e_i,r_i)}^{N}$ from $C$  
            \State \textcolor{gray}{// Knowledge Completion Phase in Equation (\ref{com_func})}
            \State {$\mathcal{T}_{(e_i,r_i)}\leftarrow M\left(I_{\tt com}\left(e_i,r_i, \mathcal{C}_{(e_i,r_i)}^{N}\right)\right)$} 
        \EndFor
        \State \textcolor{gray}{// KG Context Expansion}
        \State $\mathcal{G}_q^l\leftarrow\bigcup_{i=1}^{P}\mathcal{T}_{(e_i,r_i)}\cup\mathcal{G}_q^{l-1}$   
    \EndIf
\EndFor
\end{algorithmic}
\end{algorithm}

\para{Knowledge Completion.} Given the entity-relation pair $(e_i,r_i)$ as well as the retrieved passages $\mathcal{C}^N_{(e_i,r_i)}$,
KnowTrace further harnesses the LLM's language understanding capability to purposefully grasp key knowledge from the unstructured text.
Formally, with a completion instruction $I_{\tt com}$ that receives
$(e_i,r_i)$ and $\mathcal{C}^N_{(e_i,r_i)}$, the LLM $M$ is prompted to generate $(e_i,r_i)$-conditioned knowledge triplets:
\begin{equation}
\label{com_func}
    \mathcal{T}_{(e_i,r_i)}=M\left(I_{\tt com}\left(e_i,r_i, \mathcal{C}_{(e_i,r_i)}^{N}\right)\right).
\end{equation}
If the passages $\mathcal{C}_{(e_i,r_i)}^{N}$ do not contain relevant information to $(e_i,r_i)$, $M$ can return an empty string.
Each pair $(e_i,r_i)$ may also correspond to multiple knowledge triplets, i.e., $|\mathcal{T}_{(e_i,r_i)}|>1$, showcasing the
underlying relation mapping properties \cite{TransH, GoldE} behind the natural language text.
After completing each $(e_i,r_i)$ , KnowTrace induces a new set of knowledge triplets $\mathcal{T}=\bigcup_{i=1}^P\mathcal{T}_{(e_i,r_i)}$, offering a more comprehensive KG context $\mathcal{G}_q^l=\mathcal{G}_q^{l-1}\cup\mathcal{T}$ for the next iteration.

\para{Knowledge Prompting Strategy.} Since our KnowTrace endows the LLM
with a self-organized KG context throughout the entire inference process, one essential consideration lies in how to integrate the KG information into the LLM's prompt.
On this matter, we investigate three strategies to describe $\mathcal{G}_q$ for the prompt $I_{\tt exp}$:
\begin{itemize}[leftmargin=1.5em, itemsep=0.2em]
    \item[$\circ$] \emph{KG-to-Triplets:} directly represent $\mathcal{G}_q$ with the knowledge triplets.
    \item[$\circ$] \emph{KG-to-Paths:} connect the knowledge triplets that share common subject/object entities to form paths as the descriptions of $\mathcal{G}_q$.
    \item[$\circ$] \emph{KG-to-Texts:} leverage an additional generative model to rewrite the KG triplets into natural language texts as the descriptions.
\end{itemize}
For our framework, we demonstrate that the \emph{KG-to-Triplets} strategy offers the dual advantages of simplicity and efficacy in Section \ref{q3}.

\para{Connections to Current RAG Methods.} As shown in Figure \ref{fig:knowtrace}, 
existing iterative RAG approaches such as IRCoT \cite{IRCoT}, Self-Ask \cite{Self-Ask} and ReAct \cite{react} periodically generate new queries for retrieval, and simply accumulate all retrieved passages to form the LLM context.
In comparison, our designed framework accentuates the importance of structured thinking, which reformulates the textual accumulation process as a coherent workflow of knowledge structure expansion to empower the LLM with intelligible contexts. 
Furthermore, recent structure-enhanced methods rely on either intricate restructuring process or rigid question-parsing operations, while our KnowTrace adopts a unique perspective of structured knowledge tracing with dual merits in efficiency and flexibility. More notably, we highlight that our KnowTrace is a bootstrapping framework, which can self-distill high-quality process supervision from positive trajectories to stimulate self-taught finetuning, as described in the next section.
More discussions on the related works are included in Appendix \ref{more_discussion}.

\begin{algorithm}[t!]
\caption{Self-Bootstrapping Process of KnowTrace}\label{alg:2}
\begin{algorithmic}[1]
\Require labeled dataset $\mathcal{D}=\{(q_d,\hat{a}_d)\}_{d=1}^D$
\Input base LLM $M$
\Output bootstrapped LLM $M_K$
\State $M_0\leftarrow M$
\For{$k$ {\bfseries from} $1$ {\bfseries to} $K$}
    \State $\mathcal{D}_k\leftarrow\emptyset$
    \For{$d$ {\bfseries from} $1$ {\bfseries to} $D$}
    \State \textcolor{gray}{// Inference Process of KnowTrace (Algorithm \ref{alg:1})}
    \State $\{\mathcal{G}_{q_d},t_d,a_d\}\leftarrow\texttt{KnowTrace}(M_{k-1},q_d)$ 
    \If{$a_d==\hat{a}_d$}
    \State Collect all $\{(I_{\tt exp}(\cdot),\mathcal{P})\}$ and $\{(I_{\tt com}(\cdot),\mathcal{T}_*)\}$
    \State \textcolor{gray}{// Knowledge Backtracing (Section \ref{sft_sec})}
    \State $\mathcal{S}_{q_d}\leftarrow\texttt{Backtracing}(\mathcal{G}_{q_d},[t_d,\hat{a}_d])$    
    \State \textcolor{gray}{// Filter Out Unavailing Exploration}
    \State $\mathcal{P}^+\leftarrow\texttt{Filter}(\mathcal{P},\mathcal{S}_{q_d})$  
    \State \textcolor{gray}{// Filter Out Extraneous Completion}
    \State $\mathcal{T}^+_*\leftarrow\texttt{Filter}(\mathcal{T}_*,\mathcal{S}_{q_d})$  
    \State $\mathcal{D}_k\leftarrow\mathcal{D}_k\cup\{(I_{\tt exp}(\cdot),\mathcal{P}^+)\}\cup\{(I_{\tt com}(\cdot),\mathcal{T}^+_*)\}$
    \EndIf
\EndFor
\State \textcolor{gray}{// Finetune Base LLM on the Augmented Dataset}
\State $M_k\leftarrow\texttt{Train}(M,\mathcal{D}_k)$   
\EndFor
\State \Return $M_K$
\end{algorithmic}
\end{algorithm}

\subsection{Reflective Knowledge Backtracing for Self-Taught Finetuning}
\label{sft_sec}
Self-taught finetuning is an attractive technique, in which the LLM can post-refine its own performance without human intervention.
In line with the concept of self-training \cite{concept}, 
recent works \cite{star, rest} improve the reasoning capabilities of LLMs by training them on their own generations that ultimately lead to the correct answers.
Nevertheless, despite its effectiveness for one-time generation tasks, this process is inherently flawed when applied to
recent RAG systems in the MHQA scenario: for a complex multi-hop question, even when the final prediction is correct, a long-horizon reasoning trajectory typically still contains useless LLM generations, which would diminish the efficacy of subsequent finetuning process.

To address this limitation, the key challenge lies in how to remove the useless LLM generations and distill the contributive ones in the positive trajectories.
With the perspective of structured knowledge tracing,
our KnowTrace progressively organizes a transparent KG for the input question throughout the multi-step reasoning process.
This structured workflow allows us to design a post-hoc backtracing
mechanism to retrospectively discern the contributive generations based on the final inference.
In this way, KnowTrace can synthesize higher-quality process supervision to stimulate the self-training.

\begin{figure}[t!]
\centering
\includegraphics[width=0.95\columnwidth]{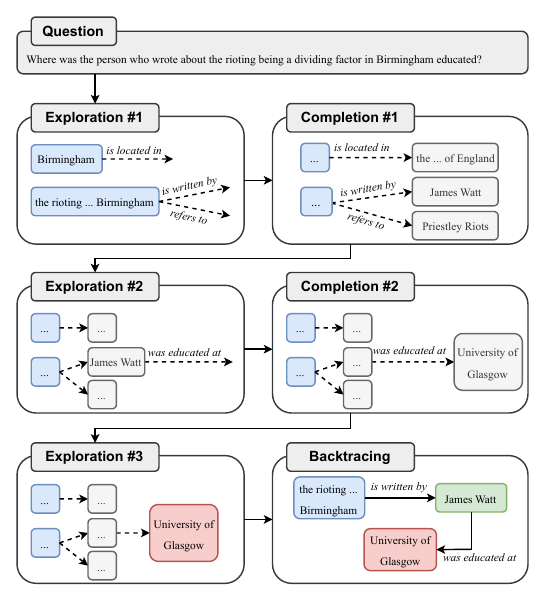}
\vspace{-1.5mm}
\caption{An example of KnowTrace's inference and backtracing process. The generated texts are included in Appendix \ref{app_case}.} \label{fig:3}
\vspace{-3mm}
\end{figure}

Formally, given a KnowTrace inference sample $(q, \mathcal{G}_{q},[t,a])$ that yields the correct answer (i.e., the prediction $a$ matches the ground-truth answer $\hat{a}$), 
the supporting knowledge essentially corresponds to a subgraph $\mathcal{S}_q\subseteq\mathcal{G}_q$ that exactly supports the final prediction.
In light of this, we adopt a simple yet effective backtracing mechanism to identify $\mathcal{S}_q$:
first, since the ground-truth label $\hat{a}$ could verify the rationality of the final outputs (i.e., $[t,a]$) \cite{CoT, R3}, we accordingly select the entities that appear in $[t,a]$ as the \emph{target entities}; then, we trace back along the graph structure
of $\mathcal{G}_q$ from these target entities to the \emph{initial entities} (defined in Section \ref{inf_sec}), thereby inducing the expected subgraph $\mathcal{S}_q$ that consists of all the supporting knowledge triplets.
Based on this subgraph, we can naturally filter out the non-contributive generations in the multi-step reasoning process:
\begin{itemize}[leftmargin=1.5em, itemsep=0.2em]
    \item[$\circ$] (\emph{Unavailing Exploration}) For the generations $\mathcal{P}$ in Equation (\ref{exp_func}), we filter out $(e_i,r_i)\in\mathcal{P}$ (or even entire $\mathcal{P}$) that fails to produce any supporting knowledge triplets in $\mathcal{S}_q$.
    \item[$\circ$] (\emph{Extraneous Completion}) For the generations $\mathcal{T}_*$ in Equation (\ref{com_func}), we filter out the completed triplets (or even entire $\mathcal{T}_*$) that do not support the final prediction (i.e., not included in $\mathcal{S}_q$).
\end{itemize}
By virtue of this process, our framework automatically removes the procedural impurities in the positive trajectories, thus synthesizing higher-quality process supervision data for self-training.
As shown in Algorithm \ref{alg:2}, KnowTrace incorporates the backtracing mechanism to bootstrap its reasoning capabilities through a simple loop:
(1) collect reasoning trajectories that lead to correct answers from a labeled MHQA dataset; (2) distill the contributive generations into a training dataset with the backtracing mechanism; (3) finetune the base LLM on this dataset; (4) restart this process to synthesize more data using the improved LLM for the next round of finetuning, until the performance plateaus.
In this way, the evident KG structures acquired by our KnowTrace not only directly enhance the inference quality, but also offer a natural way to self-synthesize high-quality supervision data for effective bootstrapping.
In Figure \ref{fig:3}, we show a specific toy example of the inference and bootstrapping procedures.

\para{Connections to Current Bootstrapping Methods.} The standard self-training approaches such as STaR \cite{star}, ReST \cite{rest} and RFT \cite{RFT} finetune LLMs on their own generations that yield correct answers. A concurrent work, InstructRAG \cite{wei2025instructrag}, directly utilizes this process to improve the one-time RAG systems.
Nevertheless, when applying this workflow to the multi-step reasoning scenarios of MHQA, its efficacy is inherently undermined by the irrelevant generations that do not contribute to the final predictions.
Our framework addresses this limitation by employing a reflective backtracing mechanism to retrospectively synthesize higher-quality supervision data, and we demonstrate that this mechanism is indispensable for the efficacy of self-taught finetuning (Section \ref{q2}). 
To the best of our knowledge, KnowTrace is the first iterative RAG framework that can effectively elevate the multi-step reasoning capabilities via self-bootstrapping.

\begin{table*}[t!]
\centering
\caption{
Evaluation results on three multi-hop question answering datasets. We adopt two advanced LLMs as the backbones, and select $N=5$ most relevant passages for each retrieval. The best results are in \textbf{bold}, and the second best results are \underline{underlined}.
}\label{tab:main}
\vspace{-1.5mm}
\begin{adjustbox}{width=0.715\textwidth,center}
\setlength\tabcolsep{6pt}
\begin{tabular}{l|cccccc|cccccc}

\toprule
& \multicolumn{6}{@{\hspace{9pt}}>{\columncolor{gray!15}}c@{\hspace{9pt}}}{\texttt{LLaMA3-8B-Instruct}} & \multicolumn{6}{|@{\hspace{9pt}}>{\columncolor{gray!15}}c@{\hspace{9pt}}}{\texttt{GPT-3.5-Turbo-Instruct}}
\\
\cmidrule(lr){2-7} \cmidrule(lr){8-13}

\multirow[c]{2}{*}{\textbf{Methods}} 
& \multicolumn{2}{c}{\textbf{HotpotQA}} 
& \multicolumn{2}{c}{\textbf{2Wiki}}
& \multicolumn{2}{c}{\textbf{MuSiQue}}
& \multicolumn{2}{|c}{\textbf{HotpotQA}} 
& \multicolumn{2}{c}{\textbf{2Wiki}}
& \multicolumn{2}{c}{\textbf{MuSiQue}}
\\
\cmidrule(lr){2-3} \cmidrule(lr){4-5} \cmidrule(lr){6-7} \cmidrule(lr){8-9} \cmidrule(lr){10-11} \cmidrule(lr){12-13}   
&  EM
&  F1
&  EM
&  F1
&  EM
&  F1
&  EM
&  F1
&  EM
&  F1
&  EM
&  F1
\\

\midrule

RA-CoT
&   .206 & .314	
&	.194 & .255
&	.132 &	.203	
&   .308 &	.429	
&	.272 & .364
&	.164 &	.258	
\\

ReAct
& .270 & .382
& .232 & .324 
& .204	& .308
& .354 & .471	
& .336 & .483
& .232 & .370	
\\

IRCoT 
& .324 & .425
& .286 & \underline{.372} 
& .240	& .332
& .442 & .565	
& .374 & .519
& \underline{.278} & \underline{.385}	
\\

Self-Ask
& .252 & .367
& .218 & .325
& .186	& .275
& .352 & .468	
& .328 & .464	
& .204 & .323
\\

Iter-RetGen
& .304 & .393
& .264 & .347
& .228 & .317
& .426 & .542	
& .368 & .495
& .246 & .372		
\\

GE-Reasoning
& .330 & .426
& .298 & .353 
& .202 & .295
& .454 & .587	
& .408 & .519
& .232 & .354	
\\

SG-Prompt
& .328 & .411
& \underline{.306} & .369 
& .236	& .342
& .448 & .583	
& .430 & .537
& .254 & .369	
\\

ERA-CoT
& \underline{.344} & \underline{.435}
& .294 & .365 
& \underline{.242} & \underline{.346}
& \underline{.460} & \underline{.592}	
& \underline{.432} & \underline{.543}
& .268 & .376	
\\

\midrule
KnowTrace
& \textbf{.386} & \textbf{.479}
& \textbf{.342} & \textbf{.403}
& \textbf{.280}	& \textbf{.387}
& \textbf{.516} & \textbf{.633}	
& \textbf{.476} & \textbf{.582}
& \textbf{.304} & \textbf{.425}	
\\

\bottomrule
\end{tabular}
\end{adjustbox}
\end{table*}

\section{Experiments}
To comprehensively demonstrate the effectiveness of our proposals, we conduct extensive experiments, which are outlined as follows:
\begin{itemize}[leftmargin=1.5em, itemsep=0.2em]
    \item[$\circ$] Firstly, we compare the basic KnowTrace with a range of RAG approaches in the MHQA task (using two mainstream LLMs as reasoning backbones), showcasing the facilitative effect of structured knowledge tracing on inference. (Section \ref{q1})
    \item[$\circ$] Secondly, we employ the designed reflective backtracing mechanism to bootstrap a new version of KnowTrace, and validate the superiority and rationality of our design from both performance and statistical perspectives. (Section \ref{q2})
    \item[$\circ$] Last but not least, we provide a detailed efficiency analysis, and investigate the effect of configuring different retrieval methods and prompting strategies on the performance. (Section \ref{q3})
\end{itemize}

\subsection{Experimental Setup}

\para{Datasets.} We evaluate our KnowTrace over three standard MHQA benchmarks under the open-domain setting:
HotpotQA \cite{HotpotQA}, 2WikiMultihopQA (2Wiki) \cite{2Wiki}, and MuSiQue \cite{trivedi2022musique}.
We use the same data splits as previous works \cite{IRCoT, sg} for evaluation.
In order to create the open-domain setting, we follow IRCoT \cite{IRCoT} to collect all candidate passages (including supporting and distractor passages) as the retrieval corpus for each dataset.
See Appendix \ref{datasets} for more details.

\para{Metrics.} We calculate the Exact Match (EM) and F1 score as metrics. The EM accuracy is the proportion of correct answers in the test set, where a prediction is deemed correct if it exactly matches one of the ground-truth answers. The F1 score evaluates the overlap between the tokens in the prediction and the answer. We apply normalization to both the predictions and the answers when computing these two metrics, following the implementation of \cite{react,sure}.

\para{Baselines.} We compare KnowTrace with a series of current RAG methods, which can be classified into three categories: 
(1) \emph{one-time retrieval-augmented} chain-of-thought reasoning \cite{CoT}, 
i.e., RA-CoT;
\\(2) \emph{iterative RAG methods}: IRCoT \cite{IRCoT}, ReAct \cite{react}, 
Self-Ask \cite{Self-Ask}, and Iter-RetGen \cite{Iter-RetGen};
(3) \emph{structure-enhanced RAG methods}: SG-Prompt \cite{sg}, GE-Reasoning \cite{[3]}, and ERA-CoT \cite{era-cot}.
Here, we describe two representative baselines that are highly relevant to our framework.
As an unstructured RAG approach, 
IRCoT \cite{IRCoT} interleaves retrieval-augmented chain-of-thought reasoning and reasoning-guided retrieval until the final answer is reported or the maximum allowed number of reasoning steps is reached.
A recent restructuring-based work, ERA-CoT \cite{era-cot}, uncover
the knowledge structures behind the textual passages with a fully LLM-driven process: first, it identifies all entities involved in the text; then, it extracts both explicit and implicit relations between entities;
next, it scores the reliability of the relations and removes those falling below a predefined threshold; after this intricate process, it performs the final answer prediction. More descriptions of all baselines can be found in Appendix \ref{baselines}.

\subsection{Implementation Details}

\para{Backbones.} We use \texttt{LLaMA3-8B-Instruct} \cite{dubey2024llama3} as the base LLM $M$ for main experiments, and also employ \texttt{GPT-3.5-Turbo-Instruct} \cite{openai2022chatgpt} to investigate the effect of distinct LLM backbones. 
The specific LLM prompts (i.e., $I_{\tt exp}$ and $I_{\tt com}$) are included in our code repository.
For each instruction, we provide four simple examples shared across all datasets to elicit the LLMs' instruction-following capabilities \cite{llm1}.
We set the temperature of $0.0$ when calling the OpenAI's API, and use greedy decoding for LLaMA, to avoid random sampling \cite{renze2024effect}.

\para{Retrievers.} Under the open-domain setting, we investigate three different retrieval models to verify the compatibility of our proposal, including
BM25 \cite{bm25}, DPR \cite{dpr}, and Contriever \cite{contriver}. 
We implement BM25 retrieval with Elasticsearch \cite{gormley2015elasticsearch}, and employ BEIR framework \cite{thakur2021beir} for DPR and Contriever.
In main experiments, we retrieve the top $N=5$ most relevant passages for each query with BM25, and also vary $N$ to $\{10, 20, 30, 50\}$ for further analysis.

\para{Self-taught finetuning.} For each dataset, we randomly sample 5,000 question-answer pairs to form $\mathcal{D}$ in Algorithm \ref{alg:2}. During the bootstrapping process, we use the designed backtracing mechanism to distill contributive generations and construct a finetuning dataset.
The statistical characteristics are analyzed in Section \ref{q2}.
On top of the base LLM, we train two distinct LoRA adapters \cite{lora} to specialize the capabilities of knowledge exploration and knowledge completion, respectively. 
We tune the training epoch in $\{1, 2, 3\}$, batch size in $\{32,64,128\}$, and learning rate
in 
$\{1\hspace{0.1mm}e\hspace{-0.6mm}-\hspace{-0.6mm}5, 5\hspace{0.2mm}e\hspace{-0.6mm}-\hspace{-0.6mm}5, 1\hspace{0.1mm}e\hspace{-0.6mm}-\hspace{-0.6mm}4, 3\hspace{0.2mm}e\hspace{-0.6mm}-\hspace{-0.6mm}4\}$. 
We repeat the bootstrapping iteration until the performance plateaus.

\begin{figure*}[t!]
\centering
\subfigure{\label{Fig-2a}
            \begin{tikzpicture}[font=\Large, scale=0.6]
                \begin{axis}[
                    legend cell align={left},
                    legend style={nodes={scale=1.0, transform shape}},
                    xlabel={Bootstrapping Iteration $k$},
                    xtick pos=left,
                    ylabel={Exact Match (EM)},
                    xmin=-0.2, xmax=4.2,
                    ymin=0.23, ymax=0.51,
                    xtick={0, 1, 2, 3, 4},
                    xticklabels={$0$, $1$, $2$, $3$, $4$},
                    ytick={0.25, 0.30,0.35,0.40,0.45,0.50},
                    yticklabels={$0.25$, $0.30$,$0.35$,$0.40$,$0.45$,$0.50$},
                    legend pos=south east,
                    ymajorgrids=true,
                    grid style=dashed,
                    title={(a) EM vs. $k$ on HotpotQA},
                    title style={yshift=-58ex, xshift=-3ex, font=\normalsize, scale=1.6},
                ]
                \addplot[
                    color=purple,
                    dotted,
                    mark options={solid},
                    mark=diamond*,
                    line width=2pt,
                    mark size=2pt
                    ]
                    coordinates {
                    (0, 0.386)
                    (1, 0.428)
                    (2, 0.440)
                    (3, 0.450)
                    (4, 0.454)
                    };
                    \addlegendentry{KnowTrace*}
                \addplot[
                    color=gray,
                    dotted,
                    mark options={solid},
                    mark=*,
                    line width=2pt,
                    mark size=2pt
                    ]
                    coordinates {
                    (0, 0.386)
                    (1, 0.392)
                    (2, 0.380)
                    (3, 0.372)
                    (4, 0.358)
                    };
                    \addlegendentry{Non-Backtracing}
                \end{axis}
                \end{tikzpicture}
    }
    \hspace{0.35mm}
    \subfigure{\label{Fig-2b}
            \begin{tikzpicture}[font=\Large, scale=0.6]
                \begin{axis}[
                    legend cell align={left},
                    legend style={nodes={scale=1.0, transform shape}},
                    xlabel={Bootstrapping Iteration $k$},
                    xtick pos=left,
                    ylabel={Exact Match (EM)},
                    xmin=-0.2, xmax=4.2,
                    ymin=0.18, ymax=0.46,
                    xtick={0, 1, 2, 3, 4},
                    xticklabels={$0$, $1$, $2$, $3$, $4$},
                    ytick={0.20, 0.25,0.30,0.35,0.40,0.45},
                    yticklabels={$0.20$, $0.25$,$0.30$,$0.35$,$0.40$,$0.45$},
                    legend pos=south east,
                    ymajorgrids=true,
                    grid style=dashed,
                    title={(b) EM vs. $k$ on 2Wiki},
                    title style={yshift=-58ex, xshift=-3ex, font=\normalsize, scale=1.6},
                ]
                \addplot[
                    color=purple,
                    dotted,
                    mark options={solid},
                    mark=diamond*,
                    line width=2pt,
                    mark size=2pt
                    ]
                    coordinates {
                    (0, 0.342)
                    (1, 0.375)
                    (2, 0.384)
                    (3, 0.402)
                    (4, 0.406)
                    };
                    \addlegendentry{KnowTrace*}
                \addplot[
                    color=gray,
                    dotted,
                    mark options={solid},
                    mark=*,
                    line width=2pt,
                    mark size=2pt
                    ]
                    coordinates {
                    (0, 0.342)
                    (1, 0.340)
                    (2, 0.340)
                    (3, 0.334)
                    (4, 0.330)
                    };
                    \addlegendentry{Non-Backtracing}
                \end{axis}
                \end{tikzpicture}
    }
    \hspace{0.35mm}
    \subfigure{\label{Fig-2c}
            \begin{tikzpicture}[font=\Large, scale=0.6]
                \begin{axis}[
                    legend cell align={left},
                    legend style={nodes={scale=1.0, transform shape}},
                    xlabel={Bootstrapping Iteration $k$},
                    xtick pos=left,
                    ylabel={Exact Match (EM)},
                    xmin=-0.2, xmax=4.2,
                    ymin=0.13, ymax=0.41,
                    xtick={0, 1, 2, 3, 4},
                    xticklabels={$0$, $1$, $2$, $3$, $4$},
                    ytick={0.15, 0.20,0.25,0.30,0.35,0.40},
                    yticklabels={$0.15$, $0.20$,$0.25$,$0.30$,$0.35$,$0.40$},
                    legend pos=south east,
                    ymajorgrids=true,
                    grid style=dashed,
                    title={(c) EM vs. $k$ on MuSiQue},
                    title style={yshift=-58ex, xshift=-3ex, font=\normalsize, scale=1.6},
                ]
                \addplot[
                    color=purple,
                    dotted,
                    mark options={solid},
                    mark=diamond*,
                    line width=2pt,
                    mark size=2pt
                    ]
                    coordinates {
                    (0, 0.280)
                    (1, 0.324)
                    (2, 0.336)
                    (3, 0.342)
                    (4, 0.342)
                    };
                    \addlegendentry{KnowTrace*}
                \addplot[
                    color=gray,
                    dotted,
                    mark options={solid},
                    mark=*,
                    line width=2pt,
                    mark size=2pt
                    ]
                    coordinates {
                    (0, 0.280)
                    (1, 0.266)
                    (2, 0.258)
                    (3, 0.252)
                    (4, 0.238)
                    };
                    \addlegendentry{Non-Backtracing}
                \end{axis}
                \end{tikzpicture}
    }

    \vspace{-1.0mm}
    \subfigure{\label{Fig-2d}
            \begin{tikzpicture}[font=\Large, scale=0.6]
                \begin{axis}[
                    legend cell align={left},
                    legend style={nodes={scale=1.0, transform shape}},
                    xlabel={Bootstrapping Iteration $k$},
                    xtick pos=left,
                    ylabel={Filtered-to-All (FA)},
                    xmin=0.8, xmax=4.2,
                    ymin=0.03, ymax=0.31,
                    xtick={1, 2, 3, 4},
                    xticklabels={$1$, $2$, $3$, $4$},
                    ytick={0.05, 0.10,0.15,0.20,0.25,0.30},
                    yticklabels={$0.05$, $0.10$,$0.15$,$0.20$,$0.25$,$0.30$},
                    legend pos=south east,
                    ymajorgrids=true,
                    grid style=dashed,
                    title={(d) FA vs. $k$ on HotpotQA},
                    title style={yshift=-58ex, xshift=-3ex, font=\normalsize, scale=1.6},
                ]
                \addplot[
                    color=purple,
                    dotted,
                    mark options={solid},
                    mark=diamond*,
                    line width=2pt,
                    mark size=2pt
                    ]
                    coordinates {
                    (1, 0.164)
                    (2, 0.135)
                    (3, 0.120)
                    (4, 0.122)
                    };
                    \addlegendentry{KnowTrace*}
                \addplot[
                    color=gray,
                    dotted,
                    mark options={solid},
                    mark=*,
                    line width=2pt,
                    mark size=2pt
                    ]
                    coordinates {
                    (1, 0.164)
                    (2, 0.180)
                    (3, 0.197)
                    (4, 0.223)
                    };
                    \addlegendentry{Non-Backtracing}
                \end{axis}
                \end{tikzpicture}
    }
    \hspace{0.35mm}
    \subfigure{\label{Fig-2e}
            \begin{tikzpicture}[font=\Large, scale=0.6]
                \begin{axis}[
                    legend cell align={left},
                    legend style={nodes={scale=1.0, transform shape}},
                    xlabel={Bootstrapping Iteration $k$},
                    xtick pos=left,
                    ylabel={Filtered-to-All (FA)},
                    xmin=0.8, xmax=4.2,
                    ymin=0.03, ymax=0.31,
                    xtick={1, 2, 3, 4},
                    xticklabels={$1$, $2$, $3$, $4$},
                    ytick={0.05, 0.10,0.15,0.20,0.25,0.30},
                    yticklabels={$0.05$, $0.10$,$0.15$,$0.20$,$0.25$,$0.30$},
                    legend pos=south east,
                    ymajorgrids=true,
                    grid style=dashed,
                    title={(e) FA vs. $k$ on 2Wiki},
                    title style={yshift=-58ex, xshift=-3ex, font=\normalsize, scale=1.6},
                ]
                \addplot[
                    color=purple,
                    dotted,
                    mark options={solid},
                    mark=diamond*,
                    line width=2pt,
                    mark size=2pt
                    ]
                    coordinates {
                    (1, 0.221)
                    (2, 0.203)
                    (3, 0.192)
                    (4, 0.165)
                    };
                    \addlegendentry{KnowTrace*}
                \addplot[
                    color=gray,
                    dotted,
                    mark options={solid},
                    mark=*,
                    line width=2pt,
                    mark size=2pt
                    ]
                    coordinates {
                    (1, 0.221)
                    (2, 0.234)
                    (3, 0.228)
                    (4, 0.242)
                    };
                    \addlegendentry{Non-Backtracing}
                \end{axis}
                \end{tikzpicture}
    }
    \hspace{0.35mm}
    \subfigure{\label{Fig-2f}
            \begin{tikzpicture}[font=\Large, scale=0.6]
                \begin{axis}[
                    legend cell align={left},
                    legend style={nodes={scale=1.0, transform shape}},
                    xlabel={Bootstrapping Iteration $k$},
                    xtick pos=left,
                    ylabel={Filtered-to-All (FA)},
                    xmin=0.8, xmax=4.2,
                    ymin=0.08, ymax=0.36,
                    xtick={1, 2, 3, 4},
                    xticklabels={$1$, $2$, $3$, $4$},
                    ytick={0.10, 0.15,0.20,0.25,0.30,0.35},
                    yticklabels={$0.10$, $0.15$,$0.20$,$0.25$,$0.30$,$0.35$},
                    legend pos=south east,
                    ymajorgrids=true,
                    grid style=dashed,
                    title={(f) FA vs. $k$ on MuSiQue},
                    title style={yshift=-58ex, xshift=-3ex, font=\normalsize, scale=1.6},
                ]
                \addplot[
                    color=purple,
                    dotted,
                    mark options={solid},
                    mark=diamond*,
                    line width=2pt,
                    mark size=2pt
                    ]
                    coordinates {
                    (1, 0.267)
                    (2, 0.238)
                    (3, 0.230)
                    (4, 0.217)
                    };
                    \addlegendentry{KnowTrace*}
                \addplot[
                    color=gray,
                    dotted,
                    mark options={solid},
                    mark=*,
                    line width=2pt,
                    mark size=2pt
                    ]
                    coordinates {
                    (1, 0.267)
                    (2, 0.274)
                    (3, 0.285)
                    (4, 0.302)
                    };
                    \addlegendentry{Non-Backtracing}
                \end{axis}
                \end{tikzpicture}
    }
    \vspace{-3mm}
    \caption{EM results (a-c) and FA ratios (d-f) in each bootstrapping iteration. KnowTrace* is the bootstrapped version based on the knowledge backtracing mechanism, and Non-Backtracing version is derived through the vanilla self-training process \cite{star}.}
    \label{fig:2}
    \vspace{-2mm}
\end{figure*}
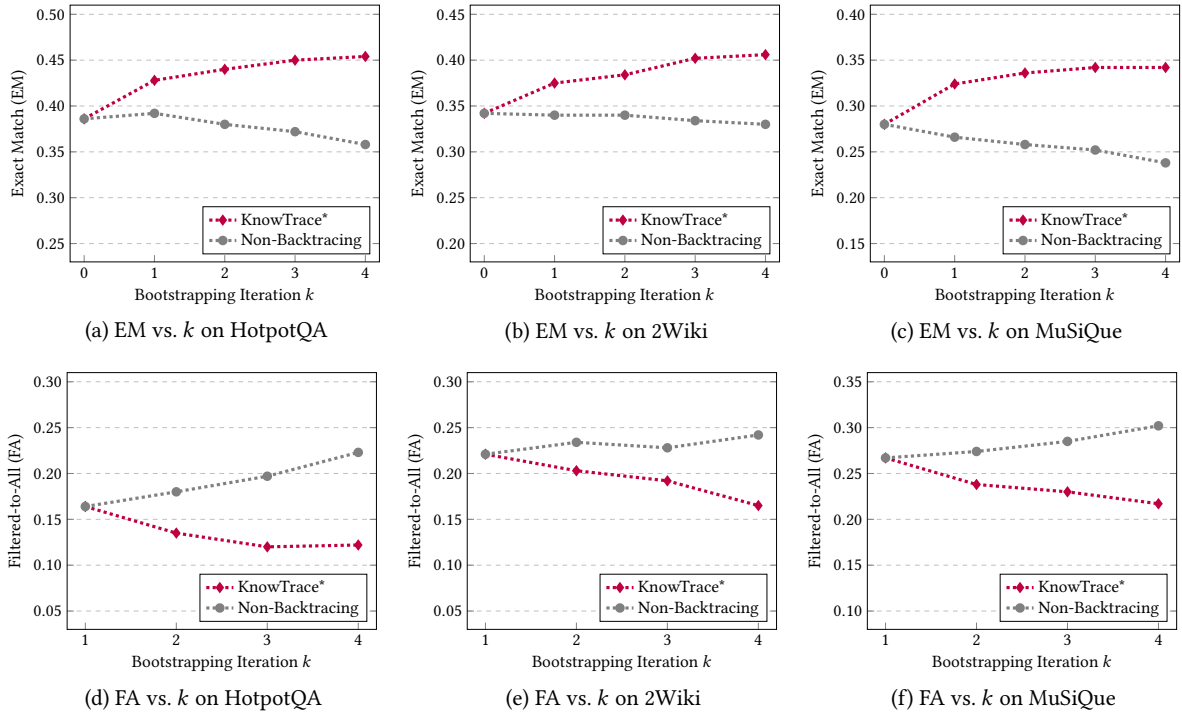

\subsection{Inference Performance Comparison}
\label{q1}
Table \ref{tab:main} summarized the main experimental results on three standard MHQA datasets.   
First, whether using \texttt{LLaMA3-8B-Instruct} or \texttt{GPT-3.5-Turbo-Instruct} as the LLM backbones, iterative RAG methods, especially IRCoT, significantly outperform the one-time RA-CoT, confirming that multi-round retrievals can indeed enhance the inference quality of the LLMs for the open-domain MHQA task.
Second, two recent restructuring-based methods, i.e., SG-Prompt and ERA-CoT, conduct one-time retrieval and strive to restructure all retrieved passages. Despite these methods retrieving only once (due to the complexity of restructuring process), they still perform comparably to or even better than the iterative methods, indicating the rationality of structured context for LLM inference.

Beyond all these methods, our framework adopts a unique perspective of structured knowledge tracing to seamlessly integrate knowledge structuring with LLM reasoning.
One can observe that KnowTrace consistently outperforms all the baselines on both evaluation metrics (i.e., EM and F1) across the three datasets.
For example, compared with IRCoT and ERA-CoT, when \texttt{LLaMA3-8B-Instruct} is selected as the base LLMs, KnowTrace achieves approximately $5.3\%$ and $4.3\%$ average absolute EM gains, respectively.
When the base LLMs are switched to \texttt{GPT-3.5-Turbo-Instruct}, the gains increase to $6.7\%$ and $4.5\%$ accordingly.
This advanced performance showcases the superiority of our design perspective at the inference level.
We further present the self-bootstrapping characteristic of our framework in the next subsection.

\subsection{Effectiveness of Knowledge Backtracing}
\label{q2}

The knowledge backtracing mechanism enables KnowTrace to synthesize a high-quality finetuning dataset for self-bootstrapping.
We refer to the bootstrapped version as KnowTrace*.
For validating its effectiveness, we follow the vanilla self-taught finetuning workflow \cite{star} to derive a \emph{non-backtracing} version as the baseline.

On the one hand, we compare their inference performance (EM) in each bootstrapping iteration.
On the other hand, we also consider such a statistical indicator: during data collection in each bootstrapping iteration, since the backtracing mechanism can naturally identify which LLM generations in the positive trajectories should be filtered, we then calculate the ratio of the tokens that should be filtered to all output tokens.
We refer to this ratio as FA (Filtered-to-All). A larger FA means that the collected positive trajectories contain more useless generations, indicating an inferior quality of the finetuning dataset.
We use this ratio to measure the proportion of noisy data in each self-training iteration.

\begin{table}[t!]
\centering
\caption{Cost statistics of KnowTrace and two baselines. \#Tok is the average count of passage tokens processed by LLMs per question; \#Time is the average inference time per question.
}
\label{tab:cost}
\vspace{-1.5mm}
\resizebox{0.95\columnwidth}{!}{
\begin{tabular}{l|cc|cc|cc}
\toprule
& \multicolumn{2}{c|}{HotpotQA} & \multicolumn{2}{c|}{2Wiki} & \multicolumn{2}{c}{MuSiQue} \\ \cmidrule(lr){2-3} \cmidrule(lr){4-5} \cmidrule(lr){6-7}
Method & \#Tok  & \#Time & \#Tok  & \#Time & \#Tok  & \#Time \\ \midrule
IRCoT & 1.4k  & 5s & 1.8k  & 7s & 2.0k  & 8s \\
ERA-CoT & 2.7k  & 16s & 3.2k  & 20s & 3.5k  & 22s \\ 
KnowTrace & 1.6k  & 6s & 1.9k  & 7s & 2.1k  & 9s \\ 
\bottomrule
\end{tabular}}
\vspace{-3mm}
\end{table}

\begin{table*}[t!]
\vspace{-0.1in}
\centering
\caption{EM results for the models using three different retrieval methods. We commonly select $N=5$ most relevant passages for each retrieval, and set \texttt{LLaMA3-8B-Instruct} as backbones.}
\label{tab:retrievers}
\vspace{-2.0mm}
\resizebox{0.75\textwidth}{!}{
\begin{tabular}{l|ccc|ccc|ccc}
\toprule
& \multicolumn{3}{c|}{HotpotQA} & \multicolumn{3}{c|}{2Wiki} & \multicolumn{3}{c}{MusiQue} \\ \cmidrule(lr){2-4} \cmidrule(lr){5-7} \cmidrule(lr){8-10}
Method & BM25 & DPR & Contriver & BM25 & DPR & Contriver & BM25 & DPR & Contriver \\ \midrule
IRCoT & .324 & .252  & .332 & .286 & .214  & .280 & .240 & .126  & .252 \\
ERA-CoT & .344 & .286  & .348 & .294 & .220  & .312 & .242 & .134  & .246 \\ 
KnowTrace & \textbf{.386} & \textbf{.320}  & \textbf{.398} & \textbf{.342} & \textbf{.246}  & \textbf{.354} & \textbf{.280} & \textbf{.176}  & \textbf{.288} \\ 
\bottomrule
\end{tabular}}
\label{table:diff_retriever}
\end{table*}

Figure \ref{fig:2} presents the EM results and FA ratios of the two versions. 
In terms of the inference performance (a-c), we observe that the backtracing-guided KnowTrace* achieves performance gains during self-training, 
while the non-backtracing version instead undergoes notable performance decline, which we attribute to its disregard for the non-contributive LLM generations in the positive trajectories.
We use the FA indicator to support this attribution from a statistical perspective (d-f).
For the fist bootstrapping iteration ($k=1$), there are more than $15\%$ (even $26.7\%$) useless generations in the collected positive trajectories. 
Indiscriminately finetuning on such noisy data results in a negative synergistic effect: the noisy data undermine the generation quality, which in turn causes the generated positive trajectories to contain more noise.
Our backtracing mechanism is able to filter out this type of noise, making it indispensable for the self-bootstrapping characteristic of our KnowTrace framework. 
We highlight that this mechanism is naturally built on the self-acquired KGs, thereby further confirming the rationality and superiority of our design perspective of structured knowledge tracing.

\subsection{More Analysis}
\label{q3}

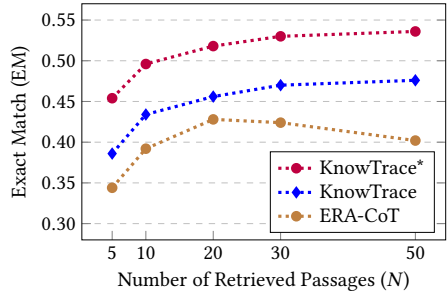
\begin{figure}[t!]
    \centering
    \vspace{1mm}
    \begin{tikzpicture}[font=\Large, scale=0.8]
                    \begin{axis}[
                        width=0.90\columnwidth, 
                        height=0.65\columnwidth,
                        legend cell align={left},
                        legend style={nodes={scale=0.9, transform shape},},
                        xlabel={Number of Retrieved Passages ($N$)},
                        xtick pos=left,
                        tick label style={font=\Large, scale=0.9},
                        ylabel style={font=\Large, scale=0.9},
                        xlabel style={font=\Large, scale=0.9},
                        ylabel={Exact Match (EM)},
                        xtick={5, 10, 20, 30, 50},
                        xticklabels={$5$, $10$, $20$, $30$, $50$},
                        ytick={0.30, 0.35,0.40,0.45,0.50,0.55, 0.60},
                        yticklabels={$0.30$, $0.35$,$0.40$,$0.45$,$0.50$,$0.55$,$0.60$},
                        ymin=0.28, ymax=0.57,
                        legend pos=south east,
                        ymajorgrids=true,
                        grid style=dashed
                    ]
                    \addplot[
                        color=purple,
                        dotted,
                        mark options={solid},
                        mark=*,
                        line width=1.5pt,
                        mark size=1.75pt
                        ]
                        coordinates {
                        (5, 0.454)
                        (10, 0.496)
                        (20, 0.518)
                        (30, 0.530)
                        (50, 0.536)
                        };
                        \addlegendentry{KnowTrace*}
                    \addplot[
                        color=blue,
                        dotted,
                        mark options={solid},
                        mark=diamond*,
                        line width=1.5pt,
                        mark size=1.75pt
                        ]
                        coordinates {
                        (5, 0.386)
                        (10, 0.434)
                        (20, 0.456)
                        (30, 0.470)
                        (50, 0.476)
                        };
                        \addlegendentry{KnowTrace}
                    \addplot[
                        color=brown,
                        dotted,
                        mark options={solid},
                        mark=*,
                        line width=1.5pt,
                        mark size=1.75pt
                        ]
                        coordinates {
                        (5, 0.344)
                        (10, 0.392)
                        (20, 0.428)
                        (30, 0.424)
                        (50, 0.402)
                        };
                        \addlegendentry{ERA-CoT}
                    \end{axis}
                    \end{tikzpicture}
                    \vspace{-3mm}
                    \captionof{figure}{EM results with varying $N$ on HotpotQA.}
                    \label{fig:N}
                    \vspace{-2.0mm}
\end{figure}

\para{Cost Analysis.} We include a detailed cost analysis for KnowTrace and two representative baselines (i.e., IRCoT and ERA-CoT). The statistics are summarized in Table \ref{tab:cost}.
In terms of the inference time and the token quantity, KnowTrace is clearly more efficient than the restructuring-based ERA-CoT, and comparable to the simplest iterative IRCoT.
Combined with the MHQA results in Table \ref{tab:main}, one can observe that our framework is capable of achieving superior effectiveness without compromising the efficiency.
Moreover, our designed backtracing mechanism does not require additional LLM calls, and automatically synthesizes high-quality data for the self-taught finetuning.
In this way, the bootstrapping cost aligns with standard parameter-efficient finetuning (approximately 2–3 hours on an NVIDIA A100 GPU).

\para{Retrieval Models.} We validate the compatibility of our framework across different retrievers.
Specifically,  in addition to BM25 used in Table \ref{tab:main}, we further conduct experiments with two other retrievers: DPR and Contriever.
Table \ref{tab:retrievers} reports the EM results of KnowTrace and two representative baselines (i.e., IRCoT and ERA-CoT) under these three retrievers. One can observe that our proposal consistently outperforms both baselines on all the datasets, regardless of the type of retrieval model. 
This superior performance demonstrates the general applicability of our approach on various retrievers.

\para{Number of Retrieved Passages.}
We further investigate the effect of the number of retrieved passages (i.e., $N$). Figure \ref{fig:N} presents the EM results of two KnowTrace versions and one restructuring-based baseline (i.e., ERA-CoT) with varying $N$ on HotpotQA dataset. From this figure, one can observe that our models consistently surpass the baseline by a clear margin across all the values of $N$.
Moreover, ERA-CoT exhibits performance degradation when $N$ is relatively large (i.e., more than $20$), which we attribute to the absence of explicit reasoning guidance during the sophisticated restructuring process. In contrast, the performance of both KnowTrace versions improves until saturation as we increase the value of $N$.
This stronger and more stable performance demonstrates the effectiveness of seamlessly integrating reasoning and structuring within our framework.

\para{Knowledge Prompting Strategies.} Since this work highlights the significance of structured knowledge structures (i.e., KGs) for LLM inference, a critical concern lies in how to incorporate the KG structures into LLM prompts. Here, we investigate three prompting strategies: \emph{KG-to-Triplets}, \emph{KG-to-Paths}, and \emph{KG-to-Texts}, respectively corresponding to elementary triplets, connected paths, and natural language texts, as described in Section \ref{inf_sec}.
Table \ref{tab:prompts} reports the EM and F1 results of our framework with these three strategies. For fair comparisons, we utilize \texttt{LLaMA3-8B-Instruct} as the base LLM and BM25 as the retriever.
One the one hand, the simplest \emph{KG-to-Triplets} works well,
while connecting independent triplets into paths (i.e., \emph{KG-to-Paths}) does not lead to consistent gains.
We observe that the path extraction process typically duplicates some triplets, which could distract LLM inference \cite{distract, hallu}.
On the other hand, converting KGs back into plain text with the LLM (i.e., \emph{KG-to-Texts}) also results in inferior performance, which we attribute to the absence of priori structural templates in the prompts. For example, when adopting the \emph{KG-to-Triplets} strategy, one can directly inform the LLM that its contexts take the triplet form of (\emph{subject}, \emph{relation}, \emph{object}).
In this way, \emph{KG-to-Triplets} exhibits the dual advantages of simplicity and effectiveness, since it avoids information duplication and also offers structural priors, making it the main choice for our framework.

\begin{table}[t!]
    \vspace{1mm}
    \caption{EM/F1 results for KnowTrace with three different knowledge prompting strategies.
    }
    \label{tab:prompts}
    \vspace{-2.0mm}
    \resizebox{0.875\columnwidth}{!}{
    \begin{tabular}{l|c|c|c}
    \toprule
    Strategy & HotpotQA & 2Wiki & MusiQue \\
    \midrule
    \emph{KG-to-Triplets} & .386/.479 & .342/.403  & .280/.387 \\
    \emph{KG-to-Paths} & .382/.465 & .334/.392  & .286/.398   \\ 
    \emph{KG-to-Texts} & .376/.471 & .320/.386  & .274/.383  \\ 
    \bottomrule
    \end{tabular}}
    \label{table:diff_prompt}
    \vspace{-3mm}
\end{table}

\section{Discussion and Conclusion}

\para{Limitations.} Despite enjoying dual merits in inference and bootstrapping, this work still has a few limitations.
First, the applicability of our design perspective in other complex scenarios, such as mathematics and decision-making tasks, has yet to be explored.
Second, although our KnowTrace can retrospectively distill high-quality data for bootstrapping, how to proactively correct erroneous trajectories without finetuning remains an open challenge.

\para{Conclusion.} This work introduces KnowTrace, an elegant iterative RAG framework that incorporates a new perspective of \emph{structured knowledge tracing} to enhance the LLM's multi-step reasoning capabilities for the MHQA task. Based on this perspective, KnowTrace  empowers the LLM with intelligible KG structures to facilitate its inference, and also employs a reflective backtracing mechanism to self-synthesize high-quality supervision data for self-bootstrapping. 
Extensive experiments on three standard MHQA benchmarks comprehensively validate the rationality and superiority of our design.

\begin{acks}
This work is supported in part by National Natural Science Foundation of China (No. 62422215 and No. 62472427), Major Innovation \& Planning Interdisciplinary Platform for the "DoubleFirst Class" Initiative, Renmin University of China, Public Computing Cloud, Renmin University of China, fund for building world-class universities (disciplines) of Renmin University of China, the Outstanding Innovative Talents Cultivation Funded Programs 2024 of Renmin University of China, and Huawei Innovation Research Programs. We gratefully acknowledge the support from Mindspore\footnote{\url{https://www.mindspore.cn}}, CANN (Compute Architecture for Neural Networks) and Ascend AI Processor used for this research.
\end{acks}

\bibliographystyle{ACM-Reference-Format}
\balance
\bibliography{sample-base}

\appendix

\section{Discussion on More Related Works}
\label{more_discussion}
This section provides a detailed discussion about more structured-enhanced works to thoroughly showcase the novelty of our design.

Several works \cite{[1],[2]} first extracts information structures from Wikipedia documents and then mask specific entities to construct a pre-training dataset, aiming to imbue the language models with structural reasoning capabilities.
In contrast, this paper focuses on enhancing the LLMs' multi-step reasoning capabilities to facilitate the MHQA task with the aid of multi-round information retrieval.

GE-Reasoning \cite{[3]} and Semi-Structured CoT \cite{[4]}
focus on parsing input questions into masked structured chains and subsequently fill each incomplete knowledge triplet based on a pre-defined KG or a plain text database.
However, such approaches heavily count on the accuracy of the initial parsing stage, thus necessitating careful filtering operations and consistency checks \cite{[4]}. In contrast, our framework adaptively traces out desired knowledge triplets throughout the multi-step reasoning process, rather than solely relying on one-time question parsing. This flexible workflow can effectively reduce error propagation and enhance robustness.

To sum up, our designed framework adopts a unique perspective of structured knowledge tracing to progressively acquire question-related knowledge triplets during the multi-step reasoning process.
This perspective enables more flexible LLM inference and does not rely on additional structural training or one-time question parsing.
Moreover, the evident KG contexts acquired during inference also provide a simple yet effective way (i.e., knowledge backtracing) to self-distill high-quality supervision data for self-taught finetuning.
Therefore, our KnowTrace framework enjoys dual advantages in the both processes of inference and bootstrapping.

\lstset{
    basicstyle=\footnotesize\ttfamily,
    breaklines=true,
    postbreak=\mbox{\hspace{-2em}{$\hookrightarrow$}\space\space}, 
    breakatwhitespace=true
}

\begin{table*}[ht!]
\centering
\small
\caption{A toy example of KnowTrace inference from the MuSiQue dataset.}
\vspace{-2.0mm}
\label{tab:example}
\renewcommand{\arraystretch}{1.0} 
\begin{adjustbox}{max width=0.8\textwidth}
\begin{tabular}{p{0.8\textwidth}}
\toprule
\textbf{Question:} Where was the person who wrote about the rioting being a dividing factor in Birmingham educated? \\
\midrule
\multicolumn{1}{c}{\textbf{\text{[}Iteration 1\text{]}}} \\
\lstset{
    basicstyle=\ttfamily\small, 
    breaklines=true,
    aboveskip=-5pt,
    belowskip=-2pt,
    breakatwhitespace=false,         
    showlines=false,                 
    numbers=none, 
}
\begin{lstlisting}
# Knowledge Memory: None
# Knowledge Exploration
Whether the collected knowledge triplets are sufficient: No
What entity-relation pairs to retrieve in the current step: 
- The rioting being a dividing factor in Birmingham: Find out who wrote about this topic and what this rioting refers to. 
- Birmingham: Find out where Birmingham is located.
# Retrieve Relevant Textual Passages From Wikipedia
# Knowledge Completion
(James Watt, wrote, the rioting being a dividing factor in Birmingham); 
(James Watt, is, an industrialist); 
(the rioting being a dividing factor in Birmingham, refers to, Priestley Riots);
(Birmingham, is located in, the West Midlands region of England)
\end{lstlisting}\\[-1mm]
\multicolumn{1}{c}{\textbf{\text{[}Iteration 2\text{]}}} \\
\lstset{
    basicstyle=\ttfamily\small, 
    breaklines=true,              
    aboveskip=-5pt,
    belowskip=-2pt,
    breakatwhitespace=false,         
    showlines=false,                 
    numbers=none, 
}
\begin{lstlisting}
# Knowledge Memory: The Knowledge Triplets Acquired in Iteration 1
# Knowledge Exploration
Whether the collected knowledge triplets are sufficient: No
What entity-relation pairs to retrieve in the current step: 
- James Watt: Find out which school James Watt attended.
# Retrieve Relevant Textual Passages From Wikipedia
# Knowledge Completion
(James Watt, was educated at, University of Glasgow)
\end{lstlisting}\\[-1mm]
\multicolumn{1}{c}{\textbf{\text{[}Iteration 3\text{]}}} \\
\lstset{
    basicstyle=\ttfamily\small, 
    breaklines=true,              
    aboveskip=-5pt,
    belowskip=-8pt,
    breakatwhitespace=false,         
    showlines=false,                 
    numbers=none, 
}
\begin{lstlisting}
# Knowledge Memory: The Knowledge Triplets Acquired in Iteration 1 and 2
# Knowledge Exploration
Whether the collected knowledge triplets are sufficient: Yes
Thought: James Watt wrote about the rioting being a dividing factor in Birmingham. He was educated at the University of Glasgow.
Answer: University of Glasgow
\end{lstlisting}\\
\bottomrule
\end{tabular}
\end{adjustbox}
\end{table*}

\section{Datasets}
\label{datasets}

\para{HotpotQA.} This is a comprehensive dataset collected from English Wikipedia, comprising approximately 113k crowd-sourced questions. The unique characteristic of HotpotQA lies in its construction, which requires answers to be derived from the introductory paragraphs of two distinct Wikipedia articles. For each question, the dataset includes the corresponding golden paragraphs from these articles, along with a curated list of sentences identified by crowdworkers as supporting evidence necessary to accurately resolve the query.
Note that our evaluation is conducted under open-domain setting \cite{IRCoT}, and thus does not use these golden information.

\para{2WikiMultihopQA (2Wiki).} This dataset consists of complex 2-hop questions that require either compositional reasoning or comparative analysis. Both structured and unstructured information from Wikipedia and Wikidata are combined for data construction. 

\para{MuSiQue.} The multi-hop questions in this dataset is constructed  by carefully selecting and composing single-hop questions obtained from a large collection of single-hop questions.
In terms of difficulty, MuSiQue is more challenging since it contains 2 to 4 hop questions.

\section{Baselines}
\label{baselines}
\para{RA-CoT \cite{CoT}.} This is the simplest approach, which conducts one-time retrieval with the input question as the query, and then utilize the retrieved passages to elicit the chain-of-thought generation.

\para{ReAct \cite{react}.} This approach integrates reasoning, action, and observation steps in an iterative process until a final answer is reached. Actions in this process include generating queries to search for relevant information or concluding with a final answer. Observations are formed by concatenating the results from these actions and serve as inputs for subsequent reasoning steps.

\para{Self-Ask \cite{Self-Ask}.} This approach follows a step-by-step workflow that breaks down the multi-hop questions and retrieve related information for each sub-question.
To improve the decomposition quality, all retrieved passages are prepended to the original question \cite{template}.

\para{Iter-RetGen \cite{Iter-RetGen}.} This approach concatenates the LLM generations from previous iterations with the original question to retrieve more relevant knowledge for the subsequent LLM inference.

\para{SG-Prompt \cite{sg}.} This approach first constructs a semantic graph structures through information extraction from all retrieved text, and then leverages this symbolic information (including entities and semantic relations) to enhance the inference quality.

\section{A Toy Example of KnowTrace}
\label{app_case}

To better illustrate the workflow of our KnowTrace framework, we present a succinct example of its inference and backtracing process, complementing the high-level presentation in Figure \ref{fig:knowtrace}.

The inference example of KnowTrace is shown in Table \ref{tab:example}. Based on the transparent KG structure traced out in this example, one can naturally trace back from the answer entity \emph{University of Glasgow} to identify the following supporting subgraph: \emph{(James Watt, wrote, the rioting being a dividing factor in Birmingham)}; \emph{(James Watt, was educated at, University of Glasgow)}. In this manner, our framework allows for removing unavailing exploration (e.g., "\emph{- Birmingham: Find out where Birmingham is located}") and extraneous completion (e.g., \emph{(James Watt, is, an industrialist)}), thereby self-distilling higher-quality process supervision data for more effective bootstrapping.

\end{document}